\documentclass[a4paper,conference]{IEEEtran}

\ifCLASSINFOpdf
\else
\fi
\usepackage{threeparttable}
\usepackage{times}
\usepackage{helvet}
\usepackage{courier}
\usepackage{color}
\usepackage{amsmath}
\usepackage{algorithm}
\usepackage{algorithmic}
\usepackage{booktabs}
\usepackage{diagbox}
\usepackage{stackengine} 
\usepackage{graphicx}
\usepackage{mathrsfs}
\usepackage{multirow}

\usepackage{epsfig,amsfonts,amssymb,times,subfigure,floatflt,wrapfig,enumerate,tikz,url,bm}
\usepackage{siunitx}
\newcommand{\mat}[1]{\bm{#1}}
\newcommand{\ten}[1]{\bm{\mathcal{#1}}} 
\DeclareMathOperator*{\argmin}{arg\,min}
\newtheorem{definition}{Definition}

\IEEEoverridecommandlockouts 
\begin{document}

% Title
\title{Exploiting Elasticity in Tensor Ranks for Compressing Neural Networks}

\author{\IEEEauthorblockN{Jie Ran*, Rui Lin*\thanks{*JR and RL contributed equally to this work.}, Hayden K.H. So, Graziano Chesi, Ngai Wong}
\IEEEauthorblockN{Department of Electrical and Electronic Engineering, The University of Hong Kong\\
Email: \{jieran, linrui, hso, chesi, nwong\}@eee.hku.hk}
}
\maketitle

% Abstract
\begin{abstract}
%Although low-rank decomposition has been widely applied in convolutional neural network (CNN) compression and achieved impressive performance, searching the optimal ranks is a crucial problem because rank controls computational complexity and accuracy in compressed models. We present a training-based rank selection method that can find elastic rank combination for a specific model. By utilizing an approximate nuclear-norm loss, the proposed method provides a good tradeoff between computational complexity and accuracy decrease. In addition, our method also reflects the correlation of ranks in different layers, and offer different level of compression. We have several observations which provide guidance on removing redundancy in convolution layers. We demonstrate the effectiveness of the proposed scheme by testing the performance of various CNNs on datasets including CIFAR-10, CIFAR-100 and ImageNet. Significant reductions in size of convolution layers are obtained, along with slight accuracy drop. 
Elasticities in depth, width, kernel size and resolution have been explored in compressing deep neural networks (DNNs). Recognizing that the kernels in a convolutional neural network (CNN) are 4-way tensors, we further exploit a new elasticity dimension along the input-output channels. Specifically, a novel nuclear-norm rank minimization factorization (NRMF) approach is proposed to dynamically and globally search for the reduced tensor ranks during training. Correlation between tensor ranks across multiple layers is revealed, and a graceful tradeoff between model size and accuracy is obtained. Experiments then show the superiority of NRMF over the previous non-elastic variational Bayesian matrix factorization (VBMF) scheme.
\end{abstract}
\IEEEpeerreviewmaketitle

% Introduction
\section{Introduction}
Deep learning and deep neural networks (DNNs) have achieved breakthroughs in various artificial intelligence (AI) applications, including classification~\cite{wu2014exploring}, object detection~\cite{liu2020cbnet}, and semantic segmentation~\cite{hong2015decoupled}. However, DNN structures are getting deeper and larger, making it challenging to deploy DNNs on edge devices with limited resources. This dilemma has motivated the search for compact neural networks with lower computation and memory cost. Various neural network compression techniques have been proposed, which can be mainly divided into three 
categories, namely, pruning~\cite{Han2016DeepCC}, quantization~\cite{cheng2017quantized, gong2014compressing}, and low-rank approximation~\cite{yu2017compressing}. %In unstructured pruning the effects lie in the sparse masks of weight matrices rather than the network architecture, so the advantages of pruned models cannot be realised without specialized hardware conditions. 

Tensor factorization~\cite{jaderberg2014speeding} belongs to the third category and is a powerful tool to compress convolutional neural networks (CNNs). It offers efficient low-rank approximations of the convolution kernels that can be regarded as a 4-way tensor, resulting in a significant reduction of parameters at the expense of only a small drop in output accuracy. Canonical polyadic (CP) decomposition is a widely used tensor decomposition method, which has been applied to decompose a 4-way kernel tensor into 
four sequential smaller convolutional (CONV) layers~\cite{lebedev2014speeding}. However, this approach is highly sensitive to the decomposition and only works well when one or two layers are compressed. Besides, the CP ranks need to be selected manually which is time-consuming without any optimality guarantee. To address this, Tucker-2 decomposition has been proposed to factorize a CONV layer into three smaller ones with Tucker ranks set by variational Bayesian matrix factorization (VBMF)~\cite{kim2015compression}. Although VBMF provides a principled way to prescribe ranks, it suffers from two major drawbacks: 1) it does not guarantee a globally or locally optimal combination of ranks; 2) once the ranks are set, they remain fixed during the fine-tuning stage, making it impossible to seek better ranks dynamically.

%Tensor factorization~\cite{jaderberg2014speeding} is a structured simplification which offers efficient representation of a convolution layer.To remove the redundancy reflected in the structural characteristics of weight matrices and improve testing efficiency on edge devices, low-rank decomposition has been proposed with different decomposition methods including matrix-based singular value decomposition (SVD) ~\cite{} and tensor-based Tucker decomposition~\cite{kim2015compression}, Canonical Polyadic (CP) decomposition~\cite{astrid2017cp}. The objective of low rank decomposition is to find a compact convolution kernel to replace the original one by setting certain constraints. The key factor in finding this similar kernel is rank, a hyperparameter which provides a tradeoff between compression ratio and accuracy decrease. Finding the optimal rank is challenging because the weight tensor decomposition in convolution layer of different neural networks have an effect on each other. Recently, Bayes-based rank selection methods including variational Bayesian matrix factorization (VBMF)~\cite{nakajima2013global} and one based Bayesian optimization~\cite{kim2020bayesian} have focused on providing theoretical optimization for rank selection, which give global solution of searching the optimal ranks. However, most of them set objectives as measuring distance between the original tensor and the decomposed one, which limit variations of weights during training and inference time. 

To improve upon VBMF rank selection, an intuitive way is to traverse all rank combinations. This is similar to the neural architecture search (NAS)~\cite{baker2016designing}, but obviously such brute-force method is time-consuming and computationally prohibitive. Recently, the once-for-all (OFA) approach has been proposed using a progressive shrinking algorithm to effectively train a network that supports diverse architectural settings with elastic depth, width, kernel size and resolution~\cite{cai2020once}. Along a related but orthogonal direction, we devise a new way to exploit elasticity in tensor ranks for compressing CNNs. Our baseline is the work in~\cite{kim2015compression} that compresses the input-output channel dimensions using Tucker-2 decomposition (the spatial dimensions, typically $3\times 3$ or $5\times 5$, are too small to be compressed). However, instead of fixing the ranks and prescribing the CNN structure before fine-tuning, our scheme dynamically finds the required Tucker-2 ranks via a nuclear-norm-like regularizer added to the normal loss function. %The final compressed model is then found by fine-tuning. 

%Using neural architecture search (NAS)~\cite{zoph2016neural} to find a specialized neural network for diverse applications has gained much attention recently. In ~\cite{cai2020once} a new method named once-for-all (OFA) with progressive shrinking algorithm was proposed to efficiently train a large number of neural networks to fit different hardware platforms and latency constraints. Inspired by OFA and considering a problem applying VBMF that searching in the neighborhood area of ranks obtained by it will frequently find better choices of ranks with higher estimation accuracy and similar compression ratio, we exploit elasticity in tensor ranks through a data-driven method.

%In this paper, we propose a new training-based approach for elastic rank selection, utilizing an auxiliary loss to restrict ranks growth during training process. Instead of going through every combination of ranks for all convolution layers and decomposing filter weight for each combination, we train a network by having a loss function with two terms for minimizing the ranks for Tucker decomposition, and we approx rank loss by minimizing the nuclear norm which is a trace function and convex. Specifically, given a trained CNN model, compressing each layer is achieved by Tucker-2 decomposition on kernel tensor. Before that, we select expected ranks by training with the added trace loss and extract singular values that preserve a certain proportion of energy to formulate the ranks. The next step is to fine-tune the network with new convolutional layers after Tucker-2 decomposition.

To our best knowledge, this is the first time that Tucker ranks of kernel tensors are found \textit{on-the-fly} during training. We further show that the ranks located by our nuclear-norm rank minimization factorization (NRMF) consistently achieve higher compression ratios than VBMF ranks with only a slight accuracy drop. Our key contributions are:

%Our experiments results show that adding a trace loss to restrict ranks can achieve higher compression ratios with a tolerable accuracy loss. Our major contributions are:

\begin{itemize}
%\item We propose a training-based strategy with a trace loss added to the loss function to exploit elastic ranks in convolution layers. In this whole compression scheme, there are three steps: (1) training a CNN with the new loss function, (2) select ranks, (3) fine-tuning.
\item We exploit the elasticity in tensor ranks during training by adding a nuclear-norm-like regularizer to the loss function, in contrast to everything being hardwired at the beginning as in the VBMF approach.

%\item {\color{red} it seems not a contribution...} By demonstrating the experiments results on various CNNs including AlexNet, VGG-16, ResNet18 and ResNet50, significant compression ratios are achieved at the small cost in accuracy.
%\item {\color{red} The description is not very clear...} 
\item By analyzing variation of ranks in early CONV layers to deeper ones, one observes an interesting decreasing of ranks in the last several layers. This could be guidance to remove redundancy in wide layers without much information loss.

\item The proposed NRMF is a generic, dynamic rank selection method which can be applied for low-rank CNN approximation together with other techniques such as quantization and pruning.
\end{itemize}

Experimental results on some popular networks demonstrate that the ranks obtained by our proposed method are better choices than VBMF-based Tucker-2 decomposition, which achieve higher compression ratios and maintain good performance. In the following, Section~\ref{sec:lit_review} shows some related works. Section~\ref{sec:tensor_base} introduces necessary tensor basics. Details of the proposed method are described in Section~\ref{sec:method}. Experiments are given in Section~\ref{sec:results} and Section~\ref{sec:conclusion} draws the conclusion.

% Lit review
\section{Related Work}
\label{sec:lit_review}

\textbf{Tensor Decomposition.}
In CNNs, computational cost during inference is dominated by the evaluation of convolutional layers. With the goal of speeding up inference, many CNN compression tools based on tensor decomposition have been proposed. Tensor-train (TT) decomposition has been applied to fully-connected layers~\cite{novikov2015tensorizing} and CONV layers~\cite{garipov2016ultimate}. %CP decomposition~\cite{astrid2017cp} has also obtained better compression results compared with Tucker-2 decomposition in~\cite{kim2015compression}. 
CP~\cite{astrid2017cp} and Tucker~\cite{kim2015compression} decomposition are also used to model compression. However, the CP approach requires manual rank setting, and is highly sensitive and only works for one or two layers.  Consequently, we use the Tucker-2 decomposition in~\cite{kim2015compression} as our baseline for compressing CONV layers.

\textbf{Rank Selection.} Ranks are key parameters in tensor decomposition for trading compression with model accuracy. In order to find the optimal ranks for different layers, several rank selection methods have been proposed. Reconstruction optimization~\cite{jaderberg2014speeding} uses CP decomposition. Ref~\cite{saha2019fitness} has proposed a heuristic fitness-based rank selection method. However, these methods have limitations in multi-rank selection and suffer from convergence problems during optimization. A sensitivity-driven rank selection considering the ratio threshold of singular values (SVs) is introduced in~\cite{chen2018spatial}, but the wide range of the criterion renders the decision difficult. Variational Bayesian matrix factorization (VBMF)~\cite{nakajima2013global} is a tool to select ranks for Tucker-2 decomposition by unfolding a tensor along different modes. VBMF offers a probabilistic ground to automatically find the ranks and noise variance, and is employed in~\cite{kim2015compression} to determine ranks. However, the matrix-wise independence in VBMF factors is hard to justify. Also, it is difficult to use VBMF to fit a given compression ratio since the parameters of controlling ranks are so sensitive that minor modifications might lead to zero ranks.

\textbf{Pruning.} The popular and effective DNN compression pipeline involves training, pruning and re-training~\cite{Han2016DeepCC}, which takes parameter magnitude for thresholding. There are various extensions: pruning cost measurement ~\cite{carreira2018learning}, soft weight-sharing and factorized Dirac posteriors~\cite{ullrich2017soft}, dynamic sparse training~\cite{liu2020dynamic} etc. Instead of pruning weights in neural networks,~\cite{cai2020once} proposes progressive shrinking to reduce multiple dimensions (viz. depth, width, kernel size and resolution) of the whole network. Besides, it targets searching the best sub-networks which maintain accuracy rather than pruning just one network. As a generalization of pruning methods, ranks used in compressing layers via Tucker decomposition can also be considered as an elastic dimension. While it has high computational cost with Tucker decomposition for every rank combination, we propose to \emph{not} actually factorize during training, but use a data-driven optimization to strike a balance between accuracy and compression.

% Preliminary
\section{Preliminary}
\label{sec:tensor_base}

\begin{figure}
\centering
\includegraphics[scale=0.25]{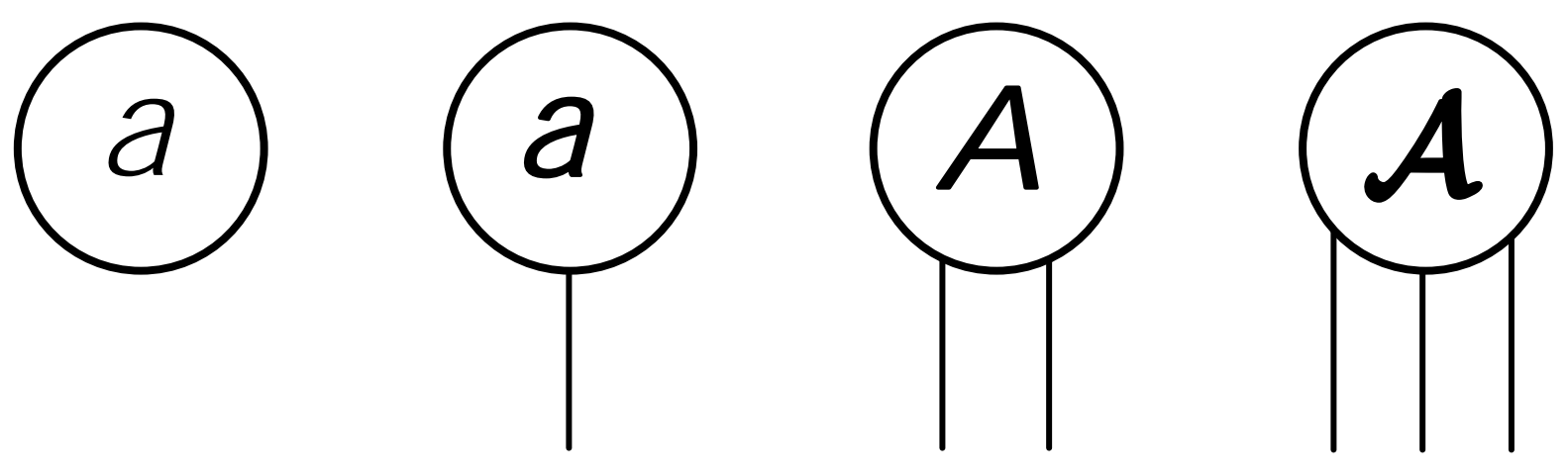}
\caption{Graphical representation of a scalar $a$, vector $\mathbf{a}$, matrix $\mathbf{A}$, and third-order tensor $\ten{A}$.}
\label{fig:TN_diagram}
\end{figure}

%Tensors are higher order generalization of vectors (viz. one-way tensors) and matrices (viz. two-way tensors)~\cite{kolda2009tensor}. We use Roman letters $a,b, \ldots$ to denote scalars; boldface capital letters $\mathbf{A,B},\ldots$ to denote matrices; and boldface capital calligraphic letters $\ten{A}, \ten{B}, \ldots$ to denote tensors. Figure~\ref{fig:TN_diagram} shows the co-called \textit{tensor network diagram} for these data structures where an open edge stands for an index axis. Using a MATLAB-like notation, ``reshape($\ten{A},[m_1,m_2,\ldots,m_p]$)'', a $d$-way tensor $\ten{A} \in \mathbb{R}^{I_1 \times I_2 \times \cdots \times I_d}$ is reshaped into another tensor with dimensions $m_1, m_2, \ldots, m_p$ that satisfy $\prod_{k=1}^p m_k=\prod_{k=1}^dI_k$. Tensor-matrix multiplication or mode product is a generalization of matrix-matrix product to that between a matrix along one mode of a tensor.

Tensors are high-dimensional generalization of vectors and matrices~\cite{kolda2009tensor}. In the following, we use Roman letters $a,b, \ldots$ to denote scalars; boldface letters $\mathbf{a,b},\ldots$ to denote vectors; boldface capital letters $\mathbf{A,B},\ldots$ to denote matrices; and boldface capital calligraphic letters $\ten{A}, \ten{B}, \ldots$ to denote tensors. \textit{Tensor network diagram} is a handy way of representing tensors as shown in Figure~\ref{fig:TN_diagram}, wherein each node denotes a tensor whose order is represented by the number of free edges. Reshaping is a basic tensor operation, by adopting a MATLAB-like notation, ``reshape($\ten{A}, [m_1, m_2, \ldots, m_p]$)'', we reshape a $d$-way tensor $\ten{A}\in \mathbb{R}^{I_1 \times I_2 \times \ldots \times I_d}$ into another $p$-way tensor with dimensions $m_1, m_2, \ldots, m_p$. The total number of elements of $\ten{A}$ is $\prod_{k=1}^dI_k$, which must be equal to $\prod_{k=1}^pm_k$. Tensor-matrix multiplication is a generalization of the matrix-matrix product to the multiplication of a matrix with a $d$-way tensor along one of its $d$ modes.

\begin{definition}(\textbf{$k$-mode product})
The $k$-mode product of tensor $\ten{G} \in \mathbb{R}^{R_1 \times \cdots \times R_d}$ with a matrix $\mat{U} \in \mathbb{R}^{J \times R_k}$ is denoted $\ten{A} = \ten{G} \times_k \mat{U}$ and defined by 
\begin{align*}
\small \ten{A}(r_1, \cdots, r_{k-1}, j, r_{k+1}, \cdots, r_d) &= \hfill \\
\small \sum_{r_k=1}^{R_k} \mat{U}(j, r_k) \ten{G}(r_1, \cdots, r_{k-1}, r_k, &r_{k+1}, \cdots, r_d)
\end{align*}
where $\ten{A} \in \mathbb{R}^{R_1 \cdots R_{k-1} \times J \times R_{k+1} \cdots \times R_d}$.
\end{definition}

\begin{definition}(\textbf{Full multilinear product})
The full multilinear product of a tensor $\ten{G} \in \mathbb{R}^{R_1 \times \cdots \times R_d}$ with matrices $\mat{U}^{(1)}, \mat{U}^{(2)}, \ldots, \mat{U}^{(d)}$, where $\mat{U}^{(k)} \in \mathbb{R}^{I_k \times R_k}$, is defined by $\ten{A} = \ten{G} \times_1 \mat{U}^{(1)} \times_2 \mat{U}^{(2)} \ldots \times_d \mat{U}^{(d)}$, where $\ten{A} \in \mathbb{R}^{I_1 \times \ldots \times I_d}$.
\end{definition}

\begin{definition}(\textbf{Tucker decomposition})
Tucker decomposition represents a $d$-way tensor $\ten{A} \in \mathbb{R}^{I_1 \times \ldots \times I_d}$ as the full multilinear product of a core tensor $\ten{G} \in \mathbb{R}^{R_1 \times R_2 \times \ldots \times R_d}$ and a set of factor matrices $\mat{U}^{(k)} \in \mathbb{R}^{I_k \times R_k}$, for $k=1,2,\ldots,d$. Writing out $\mat{U}^{(k)} = [\mat{u}_1^{(k)}, \mat{u}_2^{(k)}, \ldots, \mat{u}_{R_k}^{(k)}]$ for $k = 1,2,\ldots,d$,
\vspace{-0.45em}
\begin{align*}
\label{eqn:tucker1}
\small\ten{A}&=\small\sum\limits^{R_1}_{r_1=1}\cdots\sum\limits^{R_d}_{r_d=1}\ten{G}(r_1,\ldots,r_d)(\mat{u}_{r_1}^{(1)}\circ\cdots\circ\mat{u}_{r_d}^{(d)})\\
\small\nonumber&= \small\ten{G}\, {\times_1}\, \bm{U}^{(1)} \, {\times_2}\, \bm{U}^{(2)} \cdots {\times_d}\, \bm{U}^{(d)}
\end{align*}
where $r_1,r_2,\ldots,r_{d}$ are auxiliary indices that are summed over, and $\circ$ denotes the outer product. The dimensions $(R_1,R_2,\ldots,R_d)$ are called the \emph{Tucker ranks}.
\end{definition}

\begin{figure}
\centering
\includegraphics[scale=0.52]{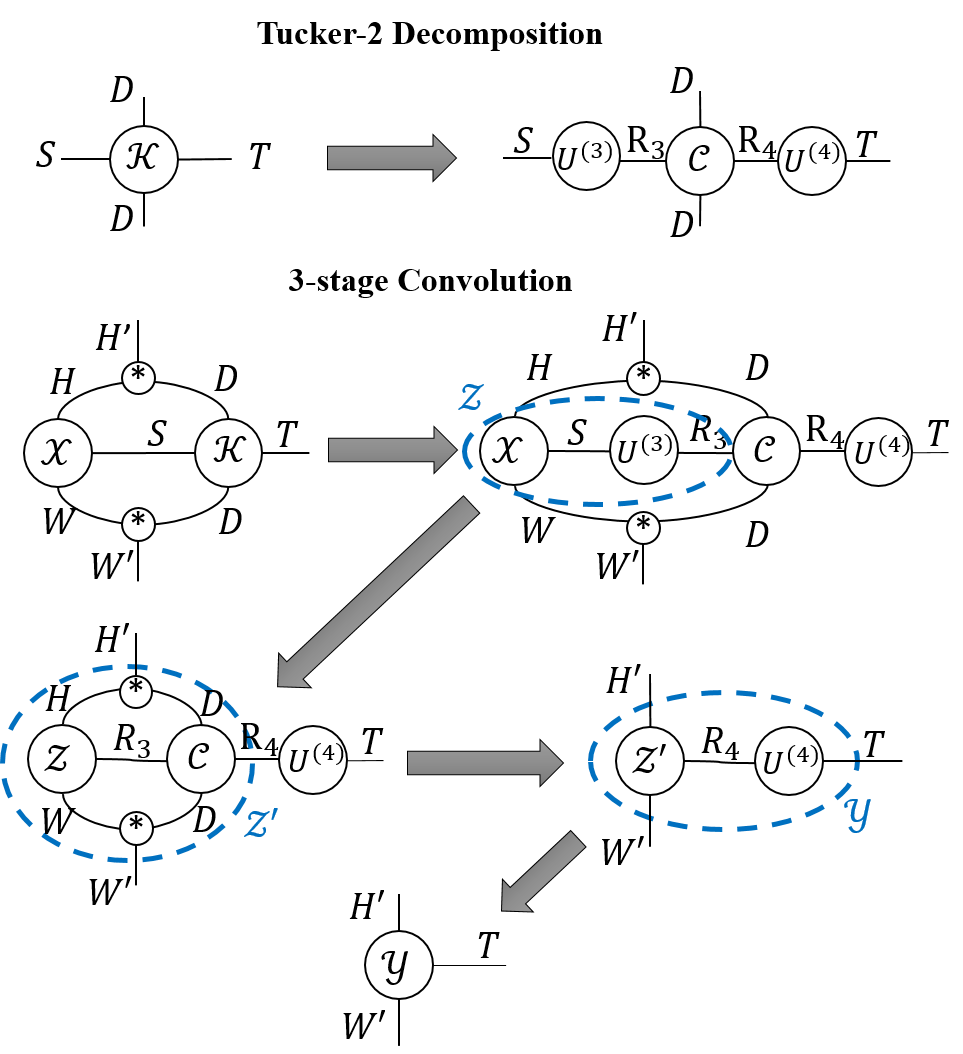}
\caption{(Upper) Tucker-2 decomposition of a kernel tensor $\ten{K}$. (Lower) Convolution process on the decomposed kernel tensor where $\ten{X}$ is the input feature and $\mat{U^{(3)}}, \ten{C}, \mat{U^{(4)}}$ are three sequential smaller CONV layers after factorization with kernel size $1\times 1$, $D \times D$ and $1 \times 1$, respectively.}
\label{fig:tucker-2}
\end{figure}

With the definition of Tucker decomposition in place, the Tucker-2 decomposition~\cite{kim2015compression} is easily understood. A CONV layer can be regarded as a $4$-way kernel tensor of size [height $\times$ width $\times$ \#input $\times$ \#outputs], and the Tucker decomposition is applied only to the last two modes instead of all four since spatial dimensions are too small (e.g., $3 \times 3$ or $5 \times 5$) to be decomposed. Figure~\ref{fig:tucker-2} shows the graphical representation of Tucker-2 decomposition on a kernel tensor $\ten{K}$ in the upper part, and the convolution process after the decomposition in the lower part. The Tucker ranks play a crucial role in approximating the original full tensor, which determines the performance of the compressed CNNs. 

\begin{figure}[t]
\centering
\includegraphics[scale=0.5]{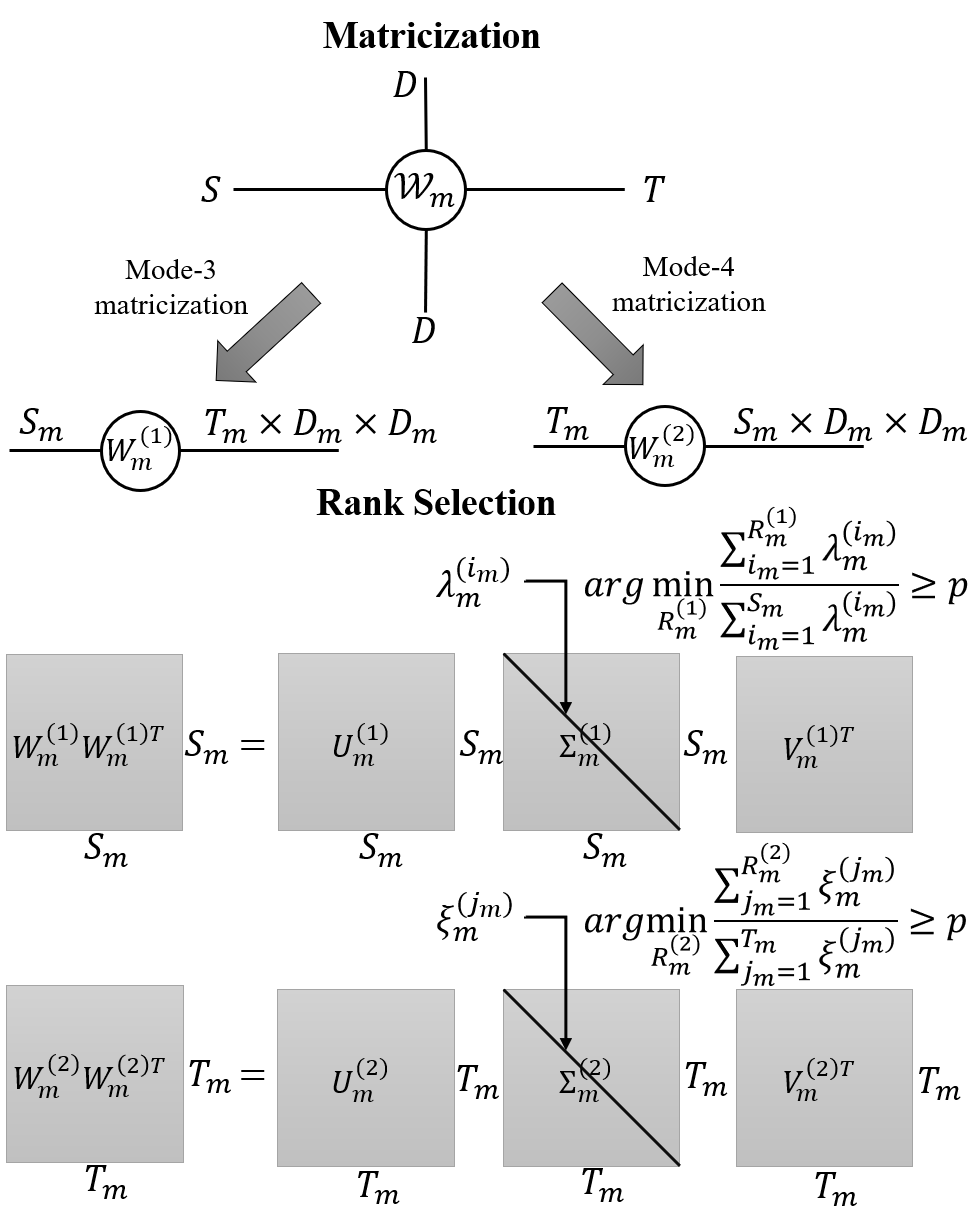}
\caption{(Upper) Mode-3 and mode-4 matricizations of $4$-way kernel tensor $\ten{W}_m$. (Lower) NRMF rank selection strategy.}
\label{fig:rank_selection}
\end{figure}

VBMF is employed in~\cite{kim2015compression} to find Tucker ranks, which we briefly describe here. The upper part of Figure~\ref{fig:rank_selection} shows mode-$3$ and mode-$4$ matricizations of the $4$-way kernel tensor $\ten{W} \in \mathbb{R}^{D \times D \times S \times T}$. When searching rank $R_3$ for $\mat{W}^{(1)} \in \mathbb{R}^{S \times (T\times D \times D)}$, VBMF treats $\mat{W}^{(1)}$ as the observation matrix, which is the sum of a target matrix $\mat{U} \in \mathbb{R}^{S\times (T \times D \times D)}$ and a noise matrix $\mat{E} \in \mathbb{R}^{S \times (T\times D \times D)}$ such that $\mat{W}^{(1)} = \mat{U} + \mat{E}$. The goal of VBMF is to find a low-rank $\mat{U}$ by filtering out the noise on $\mat{W}^{(1)}$. Being a probabilistic matrix factorization tool~\cite{nakajima2013global}, VBMF finds the noise variance on $\mat{W}^{(1)}$ automatically, then gets the rank $R_3$ of $\mat{U}$, and similarly for $R_4$.

% Methodology
\section{The Proposed NRMF Scheme}
\label{sec:method}
Here we propose a novel way to exploit elasticity in Tucker ranks. First, we introduce a nuclear-norm-based regularizer, and demonstrate how it can dynamically locate the ranks during training. Algorithm~\ref{alg:work_flow} captures the workflow.

%{\color{red} I think it is more proper to introduce our methodology in the following way.}
\subsection{Regularizer}
\label{sec:regularizer}
%{\color{red}What is the regularizer we used? how to describe the regularizer when given a neural network?}
Given a CNN with $M$ convolution filter tensors $\ten{W}_m \in \mathbb{R}^{D_m \times D_m \times S_m \times T_m}$, $m = 1, 2, \cdots, M$, where $D_m \times D_m$ denotes the kernel size, $S_m$ and $T_m$ are input and ouput channels for $m$th convolution layer, respectively. Then, we get $\mat{W}_m^{(1)} \in \mathbb{R}^{S_m \times (T_m \times D_m \times D_m)}$ and $\mat{W}_m^{(2)} \in \mathbb{R}^{T_m \times (S_m \times D_m \times D_m)}$ as mode-3 and mode-4 matricizations of $\ten{W}_m$. Next, a regularization term $L_n$ is added to the training loss that penalizes increase in the sum of singular values (SVs) of matrices $\mat{W}_m^{(1)} \mat{W}_m^{(1) T}$ and $\mat{W}_m^{(2)} \mat{W}_m^{(2) T}$:
\begin{align}
L_n = \frac{1}{2M} \sum_{m=1}^M (tr(\mat{W}_m^{(1)} \mat{W}_m^{(1) T}) + tr(\mat{W}_m^{(2)} \mat{W}_m^{(2) T})).
\end{align} 
Because $tr(\mat{W}_m^{(1)} \mat{W}_m^{(1) T})$ and $tr(\mat{W}_m^{(2)} \mat{W}_m^{(2) T})$ are also nuclear norms of $\mat{W}_m^{(1)} \mat{W}_m^{(1) T}$ and $\mat{W}_m^{(2)} \mat{W}_m^{(2) T}$, respectively, we call this trace loss a nuclear-norm loss.

\subsection{Modified Loss Function}
\label{sec:new_loss}
%{\color{red} What is the new loss function? how this new loss function help exploit the tensor ranks? How to calculate the gradients and update the parameters? how to choose the coefficient before the regularizer?}
Having the regularizer, we can train a CNN directly through back-propagation algorithm to obtain both truncated ranks and a new way of initialization which preserves compressed information from the original CNN. Given the training dataset $D = \{ (\boldsymbol{x}_1,\boldsymbol{y}_1), (\boldsymbol{x}_2,\boldsymbol{y}_2), \cdots , (\boldsymbol{x}_N,\boldsymbol{y}_N) \}$ and weights parameter $\ten{W}$ in the network, our initialization is determined by:
\begin{align}
\mathcal{J}(\ten{W}) &= \frac{1}{N}(\sum_{i = 1}^{N}\mathcal{L}((\boldsymbol{x}_i, \boldsymbol{y}_i), \ten{W})) + \alpha L_n(\ten{W}_{cl}),\\
\mat{w}^* &= \arg\min_{\ten{W}}\mathcal{J}(\ten{W}),%
\end{align}
where $\mathcal{L}$ is the loss function, e.g., cross-entropy loss in classification, $\ten{W}_{cl} \subseteq \ten{W}$ are filter weights for all convolution layers with kernel size larger than $1 \times 1$, $\alpha$ is the scaling factor for the rank regularization term. 

The regularizer $L_n$ tends to suppress non-zero SVs and yields smaller ranks. However, stronger suppression of SVs incurs increase in the loss function, which counteracts the compression level. Consequently, it becomes a game between the estimation loss term and the regularization term, through which our method can dynamically find elastic ranks that strike a balance between performance and Tucker-2 ranks.   

\subsection{Rank Selection}
\label{sec:rank_selection}
%{\color{red} After training, how can we decide the final ranks that will be used to do the Tucker-2 decomposition? How to set the threshold $p$?}
% Algorithm

%{\color{red} The following contents need to be more detailed and assigned to the above three subsections. Besides, some notations in the formulas are wrong...} 
The proposed regularizer facilitates the learning of ranks to achieve high compression ratios without much accuracy loss. After training, the SVs in $\mat{W}_m^{(1)}\mat{W}_m^{(1)T}$ and $\mat{W}_m^{(2)}\mat{W}_m^{(2)T}$ are suppressed to have nearly zero tails. By considering sum of these SVs as ``energy'' in the original weight tensor, we use a threshold ratio $p$ to retain the leading singular vectors and the number of SVs which preserve a certain percentage of energy, as detailed in Algorithm~\ref{alg:work_flow}. Because we are minimizing the sum of SVs while keeping a balance between estimation accuracy and compression ratio using threshold $p$, initialization for fine-tuning after doing Tucker-2 decomposition preserves most energy in the original network without much information loss. However, the accuracy still drops after Tucker-2 truncation, which can be recovered by fine-tuning with our Tucker-based initialization.

%{\color{red} It seems unnecessary to introduce Tucker-2 here, since it is not our contribution and the following deriviation can be found in Samsung's paper.} \textbf{Tucker-2 Decomposition} 

%{\color{red} Tucker-2 and fine-tuning can be included in Algorithm 1. Then contents in Fine-tuning can be assigned to subsection C.} \textbf{Fine-tuning} 

\begin{algorithm}[th]
\caption{Nuclear-norm rank minimizing factorization (NRMF)-based training for dynamic Tucker-2 rank finding.}
\label{alg:work_flow}
\begin{algorithmic}
% \REQUIRE pretrained CNN with new trace loss function $\mathcal{J}$, $M$ convolution filter tensors $\ten{W'}_m \in \mathbb{R}^{(D_m \times D_m \times S_m \times T_m)}$, $m = 1, 2, \cdots, M$, threshold $p$
\REQUIRE A CNN with kernel tensors $\{\ten{W}_m \in \ten{W}|\ten{W}_m \in \mathbb{R}^{D_m \times D_m \times S_m \times T_m }$, $m=1,2,\ldots,M\}$, threshold $p$, scaling coefficient $\alpha$, optimizer $\mathcal{L}$, learning rate $lr$, dataset $D$, number of training epochs $N$.
\ENSURE Tucker ranks $R_m^{(3)}, R_m^{(4)}$ with  $m = 1, 2, \ldots, M$.
\STATE NRMF-based optimizer $\mathcal{J}$ in form of Eqn.(2) $\leftarrow \mathcal{L}, \alpha, \ten{W}$
\FOR{$ i = 1,2,\ldots,N$}
\STATE $\ten{W} \leftarrow \ten{W} + lr*\frac{\partial \mathcal{J}(D,\ten{W})}{\partial \ten{W}}$
\ENDFOR
\FOR{$m = 1, 2, \cdots, M$} 
\STATE $\mat{W}_m^{(1)} \leftarrow$ reshape $(\ten{W}_m, [S_m, (T_m\times D_m\times D_m)])$ 
\STATE $\mat{W}_m^{(2)} \leftarrow$ reshape $(\ten{W}_m, [T_m, (S_m \times D_m \times D_m)])$ \\
\STATE $\lambda_m^{(i_m)} \leftarrow$ SVs for $\mat{W}_m^{(1)} \mat{W}_m^{(1)T}$, $i_m = 1, \cdots, S_m$ \\
\STATE $\xi_m^{(j_m)} \leftarrow$ SVs for $\mat{W}_m^{(2)} \mat{W}_m^{(2)T}$, $j_m = 1, \cdots, T_m$
\STATE $e_m^{(1)} \leftarrow \sum_{i_m = 1}^{S_m} \lambda_m^{(i_m)}$, $e_m^{(2)} \leftarrow \sum_{j_m = 1}^{T_m} \xi_m^{(j_m)}$ 
\STATE $R_m^{(1)} \leftarrow \argmin\limits_{\quad \quad R_m^{(1)}} \frac{\sum_{i_m = 1}^{R_m^{(1)}} \lambda_m^{(i_m)}}{e_m^{(1)}} \geq p$
\STATE $R_m^{(2)} \leftarrow \argmin\limits_{\quad \quad R_m^{(2)}}\frac{\sum_{j_m = 1}^{R_m^{(2)}} \xi_m^{(j_m)}}{e_m^{(2)}} \geq p$
\ENDFOR
\end{algorithmic}
\end{algorithm}

% Experiments
\section{Experiments}
\label{sec:results}
We evaluate our proposed rank selection strategy from three perspectives. In Section~\ref{sec:demo}, we present a simple LeNet5 example to demonstrate the SV suppression by the nuclear-norm regularizer during training. In Section~\ref{sec:4combinations}, we compare the VBMF- and NRMF-induced ranks on model initialization and compression. A layer-wise analysis of compression ratios is given in Section~\ref{sec:layerwise}. Lastly, an overview result is given in Section~\ref{sec:overall}. We implement our proposed approach for four popular networks, namely, AlexNet~\cite{krizhevsky2012imagenet}, GoogleNet~\cite{szegedy2015going}, ResNet18~\cite{he2016deep}, DenseNet~\cite{huang2017densely} on CIFAR-10~\cite{krizhevsky2009learning}, CIFAR-100~\cite{krizhevsky2009learning} and ImageNet~\cite{deng2009imagenet} to demonstrate the superiority of our method when locating ranks and performing compression. All codings are done with PyTorch and experiments run on an NVIDIA GeForce GTX1080 Ti Graphics Card with 11GB frame buffer.

%In this section, our proposed method is evaluated on CIFAR-10 and CIFAR-100~\cite{krizhevsky2009learning} with various modern network structures including AlexNet~\cite{krizhevsky2014one}, ResNet18~\cite{he2016deep} and GoogLeNet~\cite{szegedy2015going}. We evaluated the performance of our model through top-1 accuracy, top-5 accuracy and memory complexity in the decomposed CNNs. In addition, to search for a suitable threshold $p$ seeking a tradeoff between the compression ratio and prediction accuracy, we predefined three ratio threshold $p = 92\%, 95\%, 98\%$ for extracting ranks in ResNet18 on CIFAR-10. To demonstrate effectiveness of our initialization and selected ranks, we do truncated Tucker-2 decomposition using ranks by VBMF and our NRMF respectively.

%\subsection{Toy Experiment}

% SVs curves
\subsection{Effect of Regularizer on SVs of the Parameters}
\label{sec:demo}
We use a simple example to illustrate the effect of the nuclear-norm regularizer. Specifically, we apply NRMF to a modified LeNet5 on MNIST~\cite{lecun2010mnist}, namely, by inserting an extra CONV layer with a kernel tensor of size $\ten{W} \in \mathbb{R}^{3\times 3 \times 128 \times 256}$ into the original network, then training with and without the regularizer. In the test, we set the scaling coefficient $\alpha = 10^{-2}$, use a batch size of $64$ and learning rate $10^{-4}$ decaying 0.1 times every 5 epochs. We train the modified LeNet5 for 50 epochs to show the trend of SV variation.

\begin{figure*}[t]
\centering
\includegraphics[width=1\textwidth]{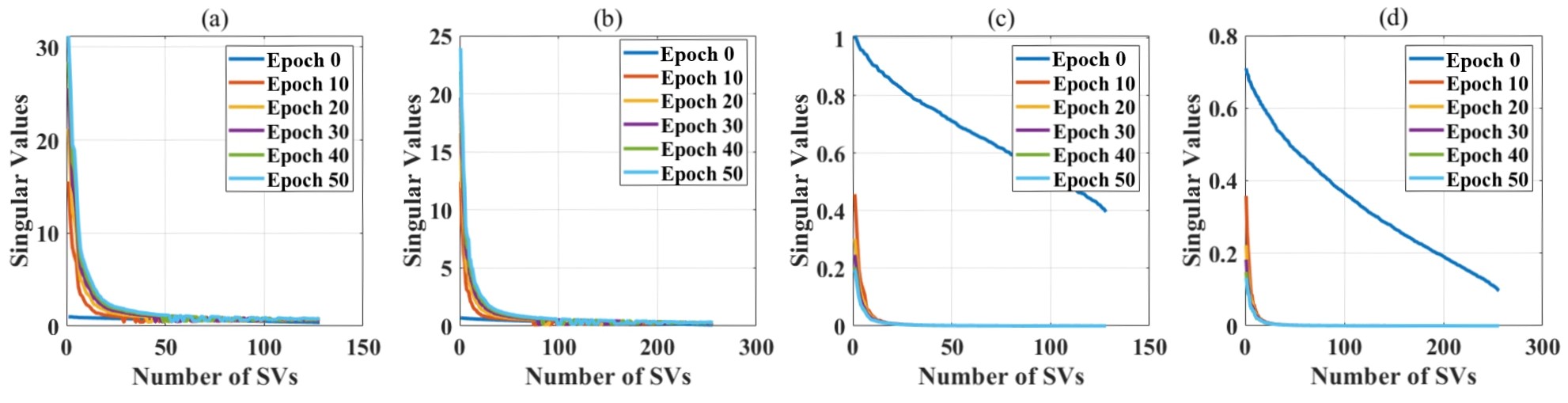}
\caption{(a) and (b) show the trends of SVs of $\mat{W}^{(1)}\mat{W}^{(1)T}$ with and without regularizer, respectively. (c) and (d) show the trends of SVs of $\mat{W}^{(2)}\mat{W}^{(2)T}$ with and without regularizer, respectively. It is obvious that when the regularizer is included in the training process, SVs keeps decreasing as the number of training epochs increases.}
\label{fig:SVs_trend}
\end{figure*}

As noted in Section~\ref{sec:tensor_base}, 
the Tucker rank selection is closely related to the SVs of $\mat{W}^{(1)} \in \mathbb{R}^{128 \times (256 \times 3 \times 3)}$ and $\mat{W}^{(2)} \in \mathbb{R}^{256 \times (128 \times 3\times 3)}$. Figures~\ref{fig:SVs_trend}(a)\&(b) show the SVs of $\mat{W}^{(1)} \mat{W}^{(1)T}$ and $\mat{W}^{(2)}\mat{W}^{(2)T}$ versus training epochs without the regularizer. It is obvious that the SVs keep increasing, meaning that the important information of $\mat{W}^{(1)}$ and $\mat{W}^{(2)}$ expands and flows through the whole network. However, as depicted in Figures~\ref{fig:SVs_trend}(c)\&(d), when the regularizer is incorporated into the training, the SVs of $\mat{W}^{(1)}\mat{W}^{(1)
T}$ and $\mat{W}^{(2)}\mat{W}^{(2)T}$ continually decrease and subside at the end. The decrease in SVs is particularly evident after the first few epochs. This phenomenon reveals that during training, the regularizer concentrates the important information flow into low-rank matrices, which facilitates subsequent model compression.

%To point out the impact of our proposed trace loss, we first conduct a toy example on a modified LeNet-5 to observe how SVs change in a convolution layer during training process. We add one convolution layer which has 128 input channels and 256 output channels on top of the original network with expanded front layers to fit the last layer. 

%In this experiment, we set the scaling coefficient $\alpha$ for trace loss as $1e-2$ and the batch size 64 for dataset MNIST. SVs are computed in mode-3 and mode-4 matricizations of weight tensor $\ten{W} \in \mathbb{R}^{3 \times 3 \times 128 \times 256}$ as $\mat{W}^{(1)} \in \mathbb{R}^{128 \times (256 \times 3 \times 3)}$ and $\mat{W}^{(2)} \in \mathbb{R}^{256 \times (128 \times 3 \times 3)}$. 

%It can be found in Fig. \ref{fig: SVs without new loss} that in training process utilizing only entropy loss for classification, singular values in unfolded matrices increase significantly during each training epoch.

%While in Fig. \ref{fig: SVs with new loss} the SVs reduce drastically from the first epoch and the largest SV in both matrices also degrade continually in training. From these curves verify that the number of SVs approaching to zero is growing which brings experimental support for rank selection using a high threshold $p$.

%\subsection{Threshold selection}
% VBMF vs. NRMF
\subsection{VBMF vs. NRMF}
\label{sec:4combinations}

% \begin{figure}[t]
% \centering
% \includegraphics[scale=0.55]{figures/workflow.png}
% \caption{Experimental procedure in Section~\ref{sec:4combinations}. Firstly, we use the pretrained model on ImageNet to train on CIFAR-10 with and without nuclear-norm regularizer. Therefore, we can obtain normal loss and nuclear loss initialization models. Next, by applying VBMF and our proposed approach to the gained models separately, we can get VBMF ranks and NRMF ranks. After this, we use the ranks to do the Tucker-2 decomposition as shown by the arrows in the figure. Therefore, we obtain VBMF and NRMF initialization models. Subsequently, 
% we use the ranks on the two models separately, and a total of four combinations, a,b,c, and d can be obtained. Lastly, we fine-tune the four models.}
% \label{fig:table345_flow}
% \end{figure}

\begin{figure*}[t]
\centering
\includegraphics[scale=0.65]{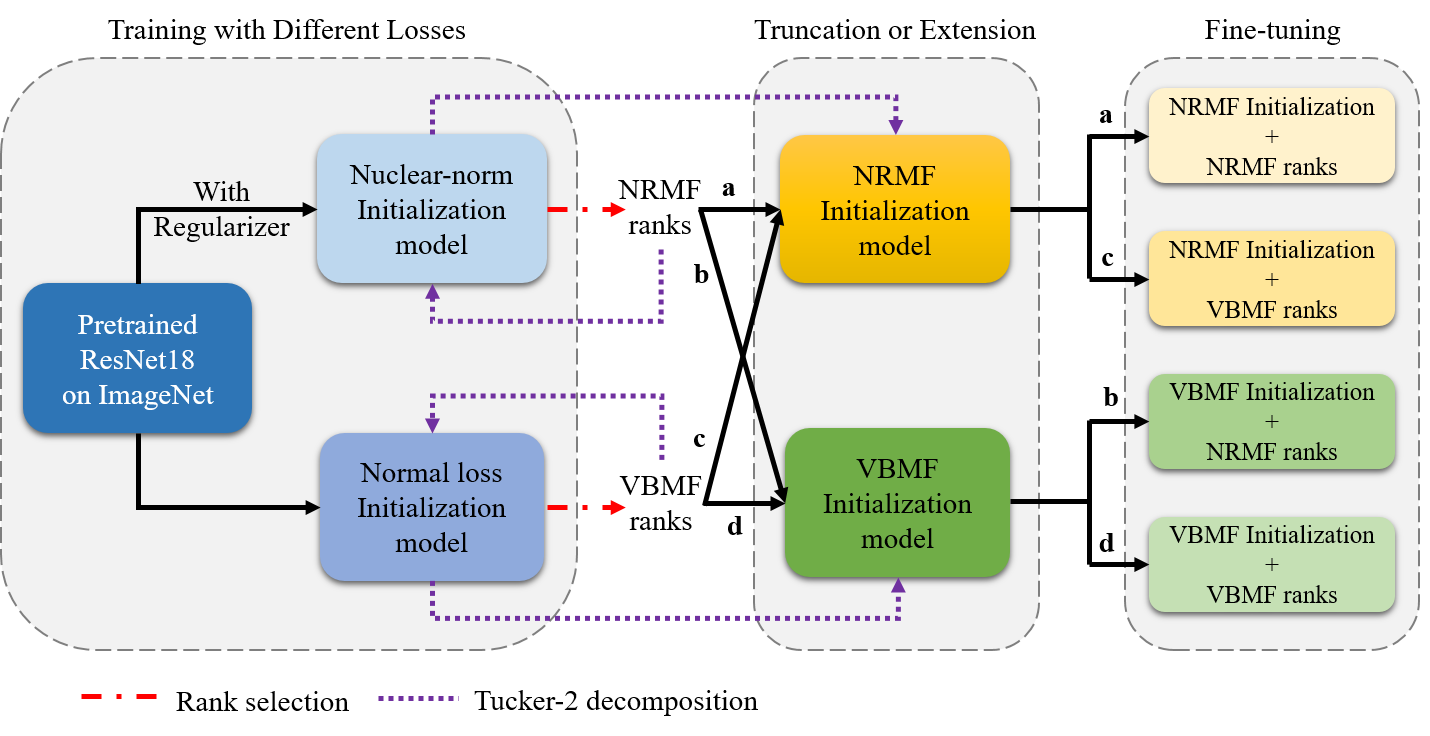}
\caption{Experimental procedure in Section~\ref{sec:4combinations}. Firstly, we use the pretrained model on ImageNet to train on CIFAR-10 with and without nuclear-norm-like regularizer. Next, by applying VBMF and NRMF to the obtained normal and nuclear-norm initialization models separately, we can get VBMF and NRMF ranks. After this, we use the ranks to do the Tucker-2 decomposition as shown by the dotted arrows. Therefore, we collect VBMF and NRMF initialization models. Next, we use each of the two sets of the ranks on the VBMF and NRMF initialization models, such that a total of four rank-initialization combinations are obtained for the fine-tuning phase.}
\label{fig:table345_flow}
\end{figure*}

% large table: p1, p2, p3
\begin{table}[t]
% p1
\renewcommand\arraystretch{1.2}
\centering
\caption{Threshold effects for ResNet18 on CIFAR-10 with $p = 92\%$}
\begin{tabular}{cccc}
\hline
& VBMF ranks & NRMF ranks  \\
\hline
VBMF initialization & $94.40\%$ & $93.50\%$\\
NRMF initialization & $95.46\%$ & $94.21\%$  \\
\hline
\# Parameters & $7.01M$ & $3.05M$ \\
\hline
\end{tabular}
\label{tab:t092}
\\[15pt]

% p2
\renewcommand\arraystretch{1.2}
\centering
\caption{Threshold effects for ResNet18 on CIFAR-10 with $p = 95\%$}
\begin{tabular}{cccc}
\hline
& VBMF ranks & NRMF ranks  \\
\hline
VBMF initialization  & $94.40\%$ & $94.40\%$ & \\
NRMF initialization & $93.58\%$ & $93.91\%$  \\ 
\hline
\# Parameters & $7.01M$ & $3.88M$\\
\hline
\end{tabular}
\label{tab:t095}
\\[15pt]

% p3
\renewcommand\arraystretch{1.2}
\centering
\caption{Threshold effects for ResNet18 on CIFAR-10 with $p = 98\%$}
\begin{tabular}{cccc}
\hline
& VBMF ranks & NRMF ranks  \\
\hline
VBMF initialization & $94.40\%$ & $95.07\%$\\
NRMF initialization & $92.98\%$ & $93.58\%$  \\
\hline
\# Parameters & $7.01M$ & $5.28M$\\
\hline
\end{tabular}
\label{tab:t098}
\end{table}

Here we compare the performances of VBMF-based and NRMF-based Tucker-2 decomposition via ResNet18 on CIFAR-10. The test setup and flow are depicted in Figure~\ref{fig:table345_flow}. As described in Section~\ref{sec:rank_selection}, the value of $p$ is important for determining the ranks. In order to explore the effect of $p$ on NRMF, we take $p=92\%,95\%,98\%$, and record the performances of NRMF under these three settings.

For clarity, here we elaborate what is done in the ``Truncation or Extension'' stage in Figure~\ref{fig:table345_flow}. Before this stage, we already have two different sets of ranks, namely, VBMF ranks and NRMF ranks. Correspondingly, we have two different initialization models. By regarding the VBMF and NRMF initialization models as the baselines, we can use VBMF and NRMF ranks to modify them. Paths~\textbf{a}\&\textbf{d} in Figure~\ref{fig:table345_flow} can be understood as entering the fine-tuning stage using the NRMF (VBMF) ranks and NRMF (VBMF) truncated Tucker-2 factors. The operations of paths~\textbf{b}\&\textbf{c} are similar, and here we take path~\textbf{b} as an example. Assume there is a CONV layer decomposed into three smaller ones of size $[1,1,128,100]$, $[3,3,100,120]$, and $[1,1,120,256]$ in VBMF initialization model, and the corresponding NRMF ranks of the given CONV layer are $110$ and $90$. Then we will modify the three CONV layers to $[1,1,128,110]$, $[3,3,110,90]$, and $[1,1,90,256]$ according to the NRMF ranks. This is done by padding zeros to the CONV layer of size $[1,1,128,100]$ and truncating the CONV layer of size $[3,3,120,256]$. As for the middle one of the three smaller CONV layers, padding and truncating operations are applied on the input and output channels, respectively. 

%Note that our added trace loss compress most of SVs of matrices unfolded along the input and output channels, we can collect the number of SVs which keep almost all features in original filters while also provide effective compression. By varying the threshold $p$ for extracting SVs, we can control compression ratio of the model generated by Tucker-2 decomposition for each convolution layer.

%In view of suggestions provided by variation curves of SVs in weight matrices unfolding, the range of threshold will not be too large in our experiment. The relationships between $p$, ranks in convolution layers and compression ratio of parameters of the network ResNet18 on CIFAR-10 are searched. 

It is shown in Tables~\ref{tab:t092} to~\ref{tab:t098} that NRMF provides higher compression ratios (viz. much smaller number of parameters) with little test accuracy loss. It is observed that the highest prediction accuracy is not achieved by the combination of NRMF ranks and initialization. One possible reason is that SVs in NRMF have been already squeezed to nearly zeros, and it is more difficult to bring them back to higher values for better accuracy via fine-tuning. We note that fine-tuning through path \textbf{b} using NRMF ranks reaches competitive precision with VBMF ranks at only half the parameters. Compared with the other two truncation ways, the NRMF initialization and ranks offer a graceful tradeoff between complexity and prediction accuracy. In the following experiments, we adopt the combination of VBMF initialization and NRMF ranks to demonstrate effectiveness of our rank selection method.

% \begin{table}[t]
% \renewcommand\arraystretch{1.2}
% \centering
% \caption{Threshold effects for ResNet18 on CIFAR-10 with $p = 92\%$}
% \begin{tabular}{cccc}
% \hline
% & VBMF ranks & NRMF ranks  \\
% \hline
% VBMF initialization & $94.40\%$ & $93.50\%$\\
% Our initialization & $95.46\%$ & $94.21\%$  \\
% \# Parameters & $7.01M$ & $3.05M$ \\
% \hline
% \end{tabular}
% \label{tab:p1}
% \end{table}

% \begin{table}[t]
% \renewcommand\arraystretch{1.2}
% \centering
% \caption{Threshold effects for ResNet18 on CIFAR-10 with $p = 95\%$}
% \begin{tabular}{cccc}
% \hline
% & VBMF ranks & NRMF ranks  \\
% \hline
% VBMF initialization & $94.40\%$ & $94.40\%$ & \\
% Our initialization & $93.58\%$ & $93.91\%$  \\ 
% \# Parameters & $7.01M$ & $3.88M$\\
% \hline
% \end{tabular}
% \label{tab:p2}
% \end{table}

% \begin{table}[t]
% \renewcommand\arraystretch{1.2}
% \centering
% \caption{Threshold effects for ResNet18 on CIFAR-10 with $p = 98\%$}
% \begin{tabular}{cccc}
% \hline
% & VBMF ranks & NRMF ranks  \\
% \hline
% VBMF initialization & $94.40\%$ & $95.07\%$\\
% Our initialization & $92.98\%$ & $93.58\%$  \\
% \# Parameters & $7.01M$ & $5.28M$\\
% \hline
% \end{tabular}
% \label{tab:p3}
% \end{table}

\begin{table}[t]
\renewcommand\arraystretch{1.5}
\centering
\caption{Layer-wise analysis on ResNet18. $S$: input channel dimension, $T$: output channel dimension, $R_3$ and $R_4$ are Tucker-2 ranks. $p = 95\%$ to select ranks.}
\begin{tabular}{cccc}
\hline
Layer & $S/R_3$ & $T/R_4$ & \#Parameters \\
\hline
conv1 & $256$ & $256$ & $589.82K$ \\
conv1 (VBMF) & $168$ & $176$ & $354.18K (\times 1.67)$ \\
conv1 (NRMF) & $144$ & $141$ & $255.70K (\times 2.31)$ \\
\hline
conv2 & $256$ & $512$ & $1.18M$ \\
conv2 (VBMF) & $194$ & $275$ & $670.61K (\times 1.76)$ \\
conv2 (NRMF) & $222$ & $299$ & $807.32K (\times 1.46)$ \\
\hline
conv3 & $512$ & $512$ & $2.36M$ \\
conv3 (VBMF) & $332$ & $328$ & $1.32M (\times 1.79)$ \\
conv3 (NRMF) & $292$ & $212$ & $815.18K (\times 2.89)$ \\
\hline
conv4 & $512$ & $512$ & $2.36M$ \\
conv4 (VBMF) & $348$ & $342$ & $1.42M (\times 1.66)$ \\
conv4 (NRMF) & $160$ & $69$ & $216.61K (\times 10.89)$ \\
\hline
conv5 & $512$ & $512$ & $2.36M$ \\
conv5 (VBMF) & $382$ & $392$ & $1.74M (\times 1.35)$ \\
conv5 (NRMF) & $31$ & $39$ & $46.72K (\times 50.50)$ \\
\hline
\end{tabular}
\label{tab:layerwise}
\end{table}

% Layer-wise analysis
\subsection{Layer-wise Analysis of Compression Ratios}
\label{sec:layerwise}

In this section, we present the layer-wise analysis of compression ratios via VBMF-based and NRMF-based Tucker-2 decomposition. We apply the two approaches to ResNet18 on CIFAR-10. 
% For NRMF training, we set the scaling coefficient $\alpha = 1e-2$, use batch size of 64, and the number of training epoch is 50. 
As for the rank selection setting, we employ $p = 95\%$. Besides, the number of epochs for fine-tuning is $50$ for both VBMF-based and NRMF-based compressed models. Furthermore, only the CONV layers with $3 \times 3$ kernels are compressed, i.e., totally $16$ CONV layers in ResNet18. We do not compress CONV layers with $1\times1$ kernels and fully connected layers. The number of parameters of those $16$ original CONV layers is $10.99M$. After compression, the number of parameters of those CONV layers becomes ${6.82M}$ and ${3.67M}$ for VBMF-based and NRMF-based compressed models, respectively. The accuracy of VBMF-based and NRMF-based compressed models are both $94.40\%$.

In Table~\ref{tab:layerwise}, the layer-wise analyses of last five compressed CONV layers are presented. It is worth noting that for conv4 and conv5, NRMF has a clear advantage over VBMF. Especially for conv5, compared with the original CONV layer, the amount of parameters is reduced by $50.5\times$. Although VBMF performs better than NRMF on conv2, the gap between their compression ratios is small and can be ignored. Overall, NRMF achieves higher compression ratios on almost every layer, and can obtain a more compact model.

These results clearly show the advantages of dynamic rank search in NRMF over the fixed-rank approach in VBMF, and also the power of NRMF in revealing the unexploited data redundancy for deeper compression.

%Furthermore, we also highlight that NRMF is more powerful than VBMF when searching ranks for CONV layers having a large amount of parameters. NRMF can find much smaller ranks than VBMF, which greatly reduce the number of parameters, while maintaining the accuracy.

%By varying the threshold in the last section, we find that ranks obtained through our method change in a regular way in all convolution layers of ResNet18 which save certain energy from the original network. As demonstrated in Table. , ranks for the matrices unfolded by weight tensors in the last two layers are dramatically low compared with original ones and those obtained by VBMF.

%In this table ranks in the last 5 convolution layers in ResNet18 on CIFAR-10 are represented because other networks including VGG-16 and ResNet50 have similar trends that wide convolution layers closer to the end have higher compression ratios. The comparison is performed on VBMF initialization with VBMF and NRMF ranks respectively, prediction is the same at $94.4\%$ while ranks through VBMF keep at about $2$ times compression ratio, our NRMF can achieve more than $200$ times at the last layer.

%We think it's an important observation since it reveals that there is much more redundancy in wider convolution layers located at the end of deep CNNs. Hence, we can compress the last layers with high rates without much information loss.

% overview results
\subsection{Performances on Various Datasets and Neural Networks}
\label{sec:overall}
Finally, NRMF is evaluated on CIFAR-10, CIFAR-100 and ImageNet with various CNNs. As before, we compare NRMF with the original networks and VBMF. We set $\alpha = 10^{-2}$ as scaling of the nuclear-norm regularizer.

\begin{table}[t]
\renewcommand\arraystretch{1.5}
\centering
\caption{Performance comparison on CIFAR-10}
\begin{tabular}{cccc}
\hline
Model & Rank Selection & Top-1 Accuracy (\%) & \#Parameters   \\
\hline
\multirow{3}*{AlexNet} & Baseline & $91.85$ & $57.04M$ \\
& VBMF & $91.29$ & $55.93M$ \\
& NRMF & $91.03$ & $55.05M$ \\
\hline
\multirow{3}*{GoogLeNet} & Baseline & $95.53$ & $5.61M$\\
& VBMF & $96.18$ & $4.20M$\\
& NRMF & $95.57$ & $4.08M$\\
\hline
\multirow{3}*{DenseNet} & Baseline & $96.56$ & $6.96M$  \\
& VBMF & $95.29$ & $5.85M$\\
& NRMF & $96.99$ & $5.85M$\\
\hline
\end{tabular}
\label{tab:cifar10}
\end{table}

\textbf{CIFAR-10 and CIFAR-100} Table~\ref{tab:cifar10} presents the compression results of Tucker-2 decomposition with VBMF and NRMF ranks for AlexNet, GoogLeNet and DenseNet on CIFAR-10. It can be seen that NRMF offers the smallest models with little difference in accuracy. Surprisingly, the NRMF-compressed DenseNet is even better than the original, reaching $96.99\%$ prediction accuracy. Table~\ref{tab:cifar100} shows the results for CIFAR-100, whereby similar observations can be made.

\begin{table}[t]
\renewcommand\arraystretch{1.5}
\centering
\caption{Performance comparison on CIFAR-100}
\setlength{\tabcolsep}{0.5mm}{
\begin{tabular}{ccccc}
\hline
Model & Rank Selection & Top-1 Acc. (\%) & Top-5 Acc. (\%) & \#Parameters   \\
\hline
\multirow{3}*{AlexNet} & Baseline & $71.12$ & $91.75$ & $57.41M$ \\
& VBMF & $69.73$ & $90.51$ & $56.32M$ \\
& NRMF & $68.97$ & $90.06$ & $55.45M$ \\
\hline
\multirow{3}*{GoogLeNet} & Baseline & $78.96$ & $95.56$ & $5.70M$\\
& VBMF & $79.50$ & $95.88$ & $4.27M$\\
& NRMF & $78.93$ & $95.25$ & $4.14M$\\
\hline
\multirow{3}*{DenseNet} & Baseline & $81.43$ & $96.30$ & $7.06M$  \\
& VBMF & $82.98$ & $96.13$ & $5.92M$\\
& NRMF & $83.53$ & $96.70$ & $5.90M$\\
\hline
\end{tabular}}
\label{tab:cifar100}
\end{table}

\begin{table}[t]
\renewcommand\arraystretch{1.5}
\centering
\caption{Performance comparison on ImageNet}
\setlength{\tabcolsep}{0.5mm}{
\begin{tabular}{ccccc}
\hline
Model & Rank Selection & Top-1 Acc. (\%) & Top-5 Acc. (\%) & \#Parameters   \\
\hline
\multirow{3}*{ResNet18} & Base & $69.76$ & $89.08$ & $11.69M$ \\
& VBMF & $67.20$ & $87.88$ & $7.50M$ \\
& NRMF & $67.27$ & $87.7$ & $6.81M$ \\
\hline
\end{tabular}}
\label{tab:imagenet}
\end{table}

\textbf{ImageNet} For ResNet18 on ImageNet (ILSVRC2012), we present performance and comparison with VBMF ranks in Table~\ref{tab:imagenet}. Again, NRMF achieves higher compression with fewer parameters and also higher top-1 accuracy. 

Additionally, we verify there is high redundancy in the last two layers in ResNet18. By varying the learning rate to $10^{-3}$ in the training process adding our new loss, we get a set of ranks with small values using $p = 95\%$. Similar to Table~\ref{tab:layerwise}, we set ranks $S$, $T$ for the last two layers conv4, conv5 as $30$, $6$ and $6$, $31$ respectively. We then do Tucker-2 decomposition on the last two convolution layers and fine-tune. The number of parameters in the compressed model is $7.01M$, with top-1 and top-5 accuracies being $67.14\%$ and $87.49\%$, respectively. Compared to compressing all convolution layers using VBMF and NRMF ranks, factorizing only two layers can also provide similar levels of compression and performance.

% Conclusion
\section{Conclusion}
We present nuclear-norm rank minimization factorization (NRMF) to exploit elasticity in the 4-way kernel tensor for CNN compression. For the first time, the ranks are dynamically found through a nuclear-norm regularizer during training, which can be perceived as a game between compression and prediction accuracy. Compared to the fixed-rank variational Bayesian matrix factorization (VBMF) approach, NRMF produces higher compression ratios in various CNN structures (viz. ResNet18, AlexNet, GoogLeNet and DenseNet). Perhaps more importantly, by observing the singular value dynamics, our scheme reveals the elasticity and redundancy patterns across CNN layers, thus providing insights and guidance in compressing specific layers.

\label{sec:conclusion}

% Reference
\bibliographystyle{IEEEtran}
\bibliography{ICPR2020}
\end{document}